\pdfoutput=1

\documentclass[11pt]{article}

\usepackage[]{ACL2023}

\usepackage{times}
\usepackage{latexsym}

\usepackage[utf8]{inputenc}

\usepackage{microtype}

\usepackage{inconsolata}
\usepackage{comment}
\usepackage{xcolor}
\usepackage{multirow}
\usepackage{graphicx}

\title{Abstractive Text Summarization Using the BRIO Training Paradigm}

\author{Khang Nhut Lam \\
  Can Tho University, Vietnam \\
 \texttt{lnkhang@ctu.edu.vn} \\\And    
  Thieu Gia Doan \\
  Can Tho University, Vietnam \\
  \texttt{dgthieu@cusc.ctu.edu.vn} 
  \AND  
  Khang Thua Pham \\
  Duy Tan University, Vietnam \\ 
  \texttt{phamthuakhang@dtu.edu.vn} \\\And  
  Jugal Kalita \\
  University of Colorado, USA \\
  \texttt{jkalita@uccs.edu} \\}

\begin{document}
\maketitle
\begin{abstract}
Summary sentences produced by abstractive summarization models may be coherent and comprehensive, but they lack control and rely heavily on reference summaries. The BRIO training paradigm assumes a non-deterministic distribution to reduce the model's dependence on reference summaries, and improve model performance during inference. This paper presents a straightforward but effective technique to improve abstractive summaries by fine-tuning  pre-trained language models, and training them with the BRIO paradigm. We build a text summarization dataset for Vietnamese, called VieSum. We perform experiments with abstractive summarization models trained with the BRIO paradigm on the CNNDM and the VieSum datasets. The results show that the models, trained on basic hardware, outperform all existing abstractive summarization models, especially for Vietnamese.  
\end{abstract}

\section{Introduction}

Text summarization reduces the size of the original text while preserving its main content. The two main approaches for constructing summaries are extractive and abstractive. Extractive summarization directly lifts sentences or words which convey key topics of the original documents, and concatenates them. Abstractive summarization discovers the primary content of the documents and generates summaries. Abstractive summaries are usually more natural and coherent than extractive summaries. 

Most abstractive summarization models follow the encoder-decoder framework. Existing abstractive summarization models are trained using maximum likelihood estimation and rely on the reference summaries. \citet{Liu2022BRIOBO} propose a BRIO training paradigm to address reliance on reference summaries by assuming non-deterministic distribution of  system-generated candidate summaries. In this paper, we use the BRIO training paradigm for abstractive summarization models to construct summaries for documents in English and Vietnamese. We make the following contributions: 
\begin{itemize}
	\item We adapt the BRIO training paradigm for abstractive summarization using BART-based and T5-based models as backbones.
	\item We present issues with the BRIO paradigm.
	\item We investigate abstractive summarization models using BARTpho-BRIO and ViT5-BRIO to obtain improved results. 
	\item We publicly release the VieSum summarization dataset for research purpose.  
\end{itemize}
The remainder of this paper is organized as follows. Related work is presented in Section 2. Section 3 introduces a large dataset for summarization in Vietnamese, named VieSum. Experiments and discussion are presented in Section 4. Section 5 concludes the paper.

\section{Related Work}
\citet{sheng2022semantic}'s Siamese Semantic Preserving Generative Adversarial Net (SSPGAN) uses a Transformer-based generator to generate summaries. A Siamese Transformer-based discriminator captures the semantic consistency between the source document and the corresponding summary. During adversarial training, the discriminator calculates a reward for each word generated. On the Gigaword dataset, SSPGAN model achieves better results than many existing abstractive text summarization models such as deep recurrent generative decoder \citep{Li2017DeepRG}, actor-critic approaches from reinforcement learning \citep{Li2018ActorCriticBT}, and Transformer~\citep{vaswani2017attention}. 

\citet{Liu2022LeveragingLI} develop the PageSum model for abstractive summarization by incorporating locality bias in both encoder and decoder. Each document is partitioned into non-overlapping pages. The encoder, which is an abstractive summarizer, encodes each page and makes local predictions. The decoder predicts output based on a weighted combination of local predictions. The authors fine-tune the BART model~\citep{Lewis2020BARTDS} for abstractive summarization and investigate several approaches to locality, such as spatial locality, discourse locality, and document locality. PageSum outperforms abstractive summarization models such as longformer encoder-decoder \citep{Beltagy2020LongformerTL}, encoder-decoder attention with head-wise positional strides \citep{Huang2021EfficientAF}, and BART with Hierarchical Attention Transformer \citep{Rohde2021HierarchicalLF}. However, PageSum  takes a long time to train, requires large memory size, and fails to capture long distance dependencies. 

Several studies use pre-trained models for abstractive text summarization. \citet{Farahani2021LeveragingPA} use mT5 \citep{Xue2021mT5AM} and sequence to sequence ParsBERT \citep{Rothe2020LeveragingPC} to construct abstractive summaries for Persian texts. T5~\citep{2020t5} and BERT~\citep{devlin2018bert} have also been used to construct abstractive summaries~\citep{Garg2021NEWSAS}. \citet{Kieuvongngam2020AutomaticTS} summarize COVID-19 biomedical research articles using BERT and GPT-2 \cite{radford2019language}. 
Features of documents are extracted and integrated into an abstractive model to improve summary generation. \citet{Nambiar2022AbstractiveSO} develop an encoder-decoder model using attention, in which POS features are incorporated to the word embedding layers to enhance the word vectors. Experiments on a dataset in Malayalam show that the integration of attention model and POS features is better than the seq2seq and attention models. \citet{Barna2022AnAA} adapt the pointer generator network for abstractive summarization by combining a pre-trained word embedding layer for transferring semantic similarity and topic features for better topic coverage. A drawback of usual abstractive summarization is the omission of named entities. To ameliorate, \citet{Berezin2022NamedEI} train a named entity recognition model based on ROBERTa to discover named entities. Then, the BART masked named entity language model is trained to pay attention on the name entities. Finally, BART is fine-tuned for text summarization.

Most studies to construct abstractive summaries in Vietnamese use an encoder-decoder framework or a pre-trained model. \citet{Quoc2019AbstractiveTS} integrate sentence positions and term frequencies into a pointer generator network with a coverage mechanism to perform the abstractive summarization for Vietnamese documents. \citet{lam2022vietnamese} construct abstractive summaries for online newspapers  using RNN with attention, BiLSTM with copy generator, standard Transformer, BERT, and sequence-to-sequence abstractive models using bottom-up approach. \citet{Phan2022ViT5PT} perform experiments to summarize Vietnamese documents using Transformer-based encoder-decoder architectures such as Transformer, PhoBERT~\citep{bartpho}, and ViT5~\citep{Phan2022ViT5PT}.

\section{VieSum Dataset}
We construct a VieSum dataset for Vietnamese consisting of 1,627,415 documents and their corresponding summaries, grouped into 23 categories. In particular, BeautifulSoup\footnote{https://www.crummy.com/software/BeautifulSoup/} and Newspaper3k\footnote{https://newspaper.readthedocs.io/en/latest/} are used to collect and extract articles from popular online newspapers in Vietnamese
such as vnexpress.net, dantri.com.vn, danviet.vn, vietnamnet.vn, laodong.vn, and vov.vn. The summaries and content documents are considered reference summaries and documents, respectively.

\section{Experimental Results} 
We perform experiments in the Google Colaboratory environment, NVIDIA Tesla T4 16GB. We use the CNNDM\footnote{https://cs.nyu.edu/~kcho/DMQA/} dataset in English, and our VieSum dataset in Vietnamese. Due to limitation of the hardware, we perform experiments with 70,000 documents picked randomly and their corresponding reference summaries from VieSum.
Each dataset is split into 3 parts including 75\% for training, 8\% for validation, and 17\% for testing.

In this paper, the pre-trained BART\textsubscript{512-length}-based and T5\textsubscript{512-length}-based models are used as backbones for generating abstractive summaries. The BART~\citep{Lewis2020BARTDS} and T5~\citep{2020t5} models are trained on the CNNDM dataset, while the BARTpho~\citep{bartpho} and ViT5~\citep{Phan2022ViT5PT} are trained on the VieSum dataset. All models are base models. To make it easy for comparison, we use the same parameters as suggested by the original authors. 

\subsection{Standard Abstractive Models}
First, we experiment and evaluate abstractive summarization approaches using standard BART-base and T5-base models.  We train the models using a batch size of 4, epoch count of 5, learning rate of $10^{-5}$, warmup\_step of 20,000, and the Adam optimizer. The results of abstractive summarization systems using the standard backbone models are presented in Table~\ref{TabStandard}.
\begin{table}
	\centering
	\begin{tabular}{llccc} \hline
\textbf{Dataset}&\textbf{System} &\textbf{R-1}&\textbf{R-2} & \textbf{R-L} \\ \hline
CNNDM&BART & 42.53& 20.21& 39.47 \\
CNNDM&T5 & 36.24&15.34&33.34\\
VieSum&BARTpho &44.59&22.57&34.60\\
VieSum&ViT5& 53.39&20.63&35.88\\ \hline
	\end{tabular} 
	\caption{ROUGE scores of abstractive summarization systems using standard backbone models.}\label{TabStandard} 
\end{table}

\subsection{Fine-tuning Abstractive Models}
To improve the quality of summaries created, we fine-tune the backbone models using the Trainer provided by Hugging Face\footnote{https://github.com/huggingface/transformers}. We do not fine-tune the BART model because it is already fine-tuned on the CNN dataset. 
Table~\ref{TabFineTunedStandard} shows the ROUGE scores of the fine-tuned abstractive models.
\begin{table}
	\centering
	\begin{tabular}{lccc} \hline
		\textbf{System} &\textbf{R-1}&\textbf{R-2} & \textbf{R-L} \\ \hline
T5 fine-tuned &41.02&19.44&38.30\\
BARTpho fine-tuned&57.94&26.56&40.83\\
ViT5 fine-tuned&57.75&26.37&40.57\\ \hline
	\end{tabular} 
	\caption{ROUGE scores of abstractive summarization systems using the fine-tuned backbone models. The T5 fine-tuned model is trained on CNNDM, while the other models are trained on VieSum.}\label{TabFineTunedStandard}
\end{table}

\subsection{Fine-tuning Abstractive Models and BRIO}
The BRIO \citep{Liu2022BRIOBO} training paradigm helps abstractive summarization models to predict tokens more accurately. \citet{Liu2022BRIOBO} use BART as the backbone model. BRIO assigns probability mass to output summary candidates based on their quality using contrastive learning. The abstractive model acts as a generation model to generate abstractive candidates in an auto-regressive way, and an evaluation model to evaluate the candidates by calculating their probability distribution. The generator is trained using the standard MLE loss, while the evaluator is trained using a contrastive loss \citep{Hadsell2006DimensionalityRB}.

In BRIO, a backbone model is used to produce $N$ abstractive summaries, the so-called \textit{candsum}s, for each document. Each \textit{candsum} is assigned a quality score by obtaining the average score of its ROUGE-1, ROUGE-2, and ROUGE-L values. In particular, \citet{Liu2022BRIOBO} use the BART\textsubscript{1024-length} model to create 16 \textit{candsum}s for each document. Next, documents, reference summaries, and corresponding \textit{candsum}s sorted by the descending quality scores are used to train the abstractive summarization model using the BRIO paradigm. We note that \citet{Liu2022BRIOBO} use the standard models as back-bones and train them with the BRIO paradigm. 

In our work, the fine-tuned backbone abstractive summarization models, presented in the previous section, are used to produce \textit{N=6} \textit{candsum}s for each document using diverse beam search \cite{vijayakumar2018diverse} with num\_beam\_groups=6, diversity\_penalty=1.0, and num\_beams=4. The abstractive summarization models are trained using a learning rate of $10^{-3}$, and the Adafactor optimizer. \citet{Liu2022BRIOBO} claim that BRIO training helps the  models reach the best performance within one epoch on the CNNDM dataset\footnote{https://github.com/yixinL7/BRIO/issues/13}. Therefore, we use one epoch for training the fine-tuned summarization models with the BRIO paradigm. 
The results of the abstractive summarization systems trained with BRIO are presented in Table~\ref{TabBRIO}.

\begin{table}
	\centering
	\begin{tabular}{lccc} \hline
		\textbf{System} &\textbf{R-1}&\textbf{R-2} & \textbf{R-L} \\ \hline
BART-BRIO& 46.40&22.47&43.00\\
T5-BRIO & 44.03&20.72&40.63\\
BARTpho-BRIO &59.12&27.01&42.05\\
ViT5-BRIO &59.50&27.33&42.76\\ \hline
	\end{tabular} 
	\caption{ROUGE scores of abstractive summarization systems, which use the fine-tuned backbone models, trained with the BRIO paradigm. BART-BRIO and T5-BRIO are trained on CNNDM, and BARTpho-BRIO and ViT5-BRIO are trained on VieSum.
	}\label{TabBRIO} 
\end{table}

\subsection{Fine-tuning Abstractive Models and BRIO-Loop}
As suggested by \citet{Liu2022BRIOBO}, we perform loop processing, using the \textit{candsum}s created by the abstractive summarization models trained with BRIO to train the models. However, after several iterations of looping, the ROUGE scores seem to change very little. Especially, BARTpho and ViT5 almost reach the highest ROUGE scores with 2 iterations. Table~\ref{TabFintuneBRIO} presents the ROUGE scores obtained after looping twice.

\begin{table}
	\centering
	\begin{tabular}{lccc} \hline
		\textbf{System} &\textbf{R-1}&\textbf{R-2} & \textbf{R-L} \\ \hline
BART-BRIO-Loop  &46.55&22.56&43.00\\
T5-BRIO-Loop  &45.24&21.50&41.80\\
BARTpho-BRIO-Loop &60.53&28.20&44.20\\
ViT5-BRIO-Loop &60.90&28.39&44.36\\ \hline
	\end{tabular} 
	\caption{ROUGE scores of abstractive summarization systems trained with the BRIO paradigm after looping twice. BART-BRIO and T5-BRIO are trained on CNNDM, and BARTpho-BRIO and ViT5-BRIO are trained on VieSum.
	}\label{TabFintuneBRIO} 
\end{table}

Experimental results show that the BRIO training paradigm significantly helps improve the abstractive summaries by reducing the dependence of the system on the reference summaries. However, assigning weights to both \textit{candsum}s and reference summaries is necessary in order to decrease reliance on reference summaries. The diverse beam search helps obtain diverse \textit{candsum}s, but could cause interference in the beam search space because the model might not follow the reference summaries. In addition, using the ROUGE metric for evaluating the abstractive summarization models trained with the BRIO paradigm seems unfair because these models could produce summaries which are independent on the reference summaries.

\subsection{Discussion}

It is not easy to make comparisons between models trained on different hardware and on different datasets. We make an attempt to compare our work with published papers on similar datasets.

Curently, BRIO using a standard BART\textsubscript{1024-length} model as  backbone, which generates 16 \textit{candsum}s, achieves SOTA results on the CNNDM dataset with a ROUGE-1 of 47.78 and a ROUGE-L of 32.58 \citep{Liu2022BRIOBO}. In addition, BART\textsubscript{1024-length}-BRIO with 2 iterations reaches ROUGE-1 and ROUGE-L of 48.01 and 44.67, respectively; these are both better than our BART\textsubscript{512-length}-BRIO, which creates 6 \textit{candsum}s for each document, after 2 iterations: 46.55 for ROUGE-1 and 43.00 for ROUGE-L.

\citet{T2022ComparativeAO} fine-tune the T5 abstractive summarization model and evaluate on the CNNDM dataset. Their T5 model achieves ROUGE-1 and ROUGE-L scores of 40.79 and 34.80, respectively, which are lower than the scores of our fine-tuned T5 model, and significantly lower than scores of our best model, the T5-BRIO-Loop model: 45.24 for ROUGE-1 and 41.80 for ROUGE-L.

For Vietnamese abstractive summarization, \citet{Quoc2019AbstractiveTS} use LSTMs with the features of sentence positions and term frequencies (LSTM+SP+TF) on a Vietnamese dataset collected from Baomoi\footnote{https://baomoi.com/}. The best ROUGE-1 and ROUGE-L scores of their model are 31.89 and 29.97, respectively, which are significantly lower than the scores of our BRIO-BART model. 

Both the BARTpho and ViT5 models trained with the BRIO paradigm outperform all models proposed by~\citet{lam2022vietnamese} on the CTUNLPSum dataset, which is very similar to the VieSum dataset, including the sequence-to-sequence models, copy generator network, sequence-to-sequence with re-writer approach, and  bottom-up approach. 

\citet{bartpho} apply several models for abstractive summarization on the VNDS \citep{Nguyen2019VNDSAV} dataset. They perform experiments on 8 A100 GPUs with 40GB each. Their model is trained for 15 epochs in about 6 days. Their best model, BARTpho, achieves a ROUGE-1 of 61.14, which is slightly higher than the BARTpho-BRIO-Loop, and a ROUGE-L of 40.15, which is lower than that of the BARTpho-BRIO-Loop. In addition, the BARTpho-BRIO-Loop is trained on one epoch in about 32 hours using basic hardware.

\citet{Phan2022ViT5PT} introduce a pre-trained text-to-text Transformer for Vietnamese abstractive summarization, called ViT5. The authors claim the ViT5 model as the SOTA for Vietnamese abstractive summarization. Their ViT5 abstractive summarization model achieves ROUGE-1 and ROUGE-L of 61.85 and 41.70, respectively, on the VNDS dataset \citep{Nguyen2019VNDSAV}. We conducted experiments on VNDS and found interesting results related to the ViT5 model. The ROUGE scores of the ViT5 model trained using the common paradigm are essentially identical to the ROUGE scores provided by \citet{Phan2022ViT5PT}. However, the scores of the ViT5 model trained using the BRIO paradigm are reduced to 59.37 and 41.6, respectively. On the VieSum dataset, the standard ViT5-base achieves an ROUGE-1 of 53.39 and ROUGE-L of 35.88; while the ViT5-BRIO-Loop has better scores: ROUGE-1 of 60.90 and ROUGE-L of 44.36. We leave further exploration and  evaluation these unstable results for future work.

\section{Conclusion}
We investigated abstractive summarization models trained with the BRIO paradigm. Experiments show that we can improve abstractive summarization models by fine-tuning the backbones before training them with BRIO. In particular, the summarization models trained with BRIO outperform other summarization models in Vietnamese. We also discuss issues with the BRIO paradigm for further exploration. In addition, we built the VieSum dataset for summarization in Vietnamese. For future work, we will ask volunteers to evaluate and provide feedback on a small subset of the VieSum dataset.

\section*{Limitations}
While many studies show that the architectures of the deep learning models significantly influence the results, we perform experiments with several base architectures because of the constrained hardware. Furthermore, there has not been a Vietnamese benchmark summarization dataset, which is both sizable and of high quality. The existing summarization datasets are derived from online magazines, which usually contain misspelled words and grammatical errors. In addition, the reference summaries might not convey the main content of the corresponding articles. Therefore, selecting and developing efficient summarization models for Vietnamese still present numerous challenges.

\section*{Ethics Statement}
 We use several different software tools in our experiments. These tools as well the English dataset are publicly available and we do not see any ethical issues in using them. In addition, we clearly reference the papers and other sources for the tools used. We create the VieSum dataset ourselves. 
 
 Our paper's work depends on using previously published approaches to abstractive summarization. We clearly give credit to the authors of these approaches by citing original sources. 
 
 This paper focuses on abstractive summarization of longer documents. There is potential for high quality abstractive summarizers to be misused. For example, students if/when given an assignment to summarize/review papers/articles may use such summarizers to automatically write reviews and claim them as their own. However, we believe abstractive summarizers for long documents have not achieved this level of sophistication at this time.

\bibliography{anthology,custom}
\bibliographystyle{acl_natbib}

\end{document}